\pdfoutput=1
\documentclass[twocolumn,preprintnumbers,amsmath,amssymb,letter]{revtex4}
\pdfoutput=1
\usepackage{graphicx}
\usepackage{dcolumn}
\usepackage{bm}
\newcommand{\ra}{\rangle}
\newcommand{\la}{\langle}
\newcommand{\ttt}{\texttt}

\begin{document}

%\preprint{LAUR-00-0000}

\title{Modeling Computations in a Semantic Network}

\author{Marko A. Rodriguez}
%\email{marko@lanl.gov}
\affiliation{Digital Library Research and Prototyping Team \\
		Los Alamos National Laboratory \\
		Los Alamos, New Mexico 87545 }

\author{Johan Bollen}
%\email{jbollen@lanl.gov}
\affiliation{Digital Library Research and Prototyping Team \\
		Los Alamos National Laboratory \\
		Los Alamos, New Mexico 87545 }

\date{\today}
\begin{abstract}
Semantic network research has seen a resurgence from its early history in the cognitive sciences with the inception of the Semantic Web initiative. The Semantic Web effort has brought forth an array of technologies that support the encoding, storage, and querying of the semantic network data structure at the world stage. Currently, the popular conception of the Semantic Web is that of a data modeling medium where real and conceptual entities are related in semantically meaningful ways. However, new models have emerged that explicitly encode procedural information within the semantic network substrate. With these new technologies, the Semantic Web has evolved from a data modeling medium to a computational medium. This article provides a classification of existing computational modeling efforts and the requirements of supporting technologies that will aid in the further growth of this burgeoning domain.\\ \\
\textbf{keywords:} I.2.12 Intelligent Web Services and Semantic Web - I.2.4.k Semantic networks - I.2.4 Knowledge Representation Formalisms and Methods - I.2 Artificial Intelligence - I Computing Methodologies
\end{abstract}

\maketitle{}

\section{Introduction}

A semantic network is generally defined by a directed labeled graph \cite{know:sowa1999}. Formally, a directed labeled graph can be represented in set theoretic notation as $G = (V, E \subseteq V \times V, \lambda: E \rightarrow \Sigma)$, where $V$ is the set of vertices, $E$ is the set of edges, and $\lambda$ is a function that maps the edges in $E$ to the set of labels in $\Sigma$. Another perspective would organize each label type according to its own edge group and in such cases, $G = (V, \mathbb{E} = \{E_0, E_1, \ldots, E_n\})$, where $\mathbb{E}$ is the set of all labeled edge sets, $E_i \in \mathbb{E}$ is a particular labeled edge set, and $E_i \subseteq V \times V$ \cite{netanal:brandes2005}.

For the Semantic Web, the semantic network substrate is defined by the constraints of the Resource Description Framework (RDF) \cite{spinweb:fensel2003,rdfspec:manola2004}.  RDF represents a semantic network as a set of triples where both vertices and edge labels are called resources. In RDF, a subject resource ($s$) points to an object resource ($o$) according to a predicate resource ($p$). Subject and predicate resources are identified by Uniform Resource Identifiers (URI) \cite{uri:2001} and the object is either a literal or a URI. If $U$ is the set of all URIs and $L$ is the set of all literals, then the Semantic Web can be formally defined as $G \subseteq (U \times U \times (U \cup L))$. This representation is called a triple list where a triple $\tau = \la s, p, o \ra$. RDF is a framework (or model) for denoting a semantic network in terms of URIs and literals. RDF is not tied to a particular syntax. Various RDF syntaxes have been developed to support the encoding and distribution of RDF graphs \cite{rdfsyntax:beckett2003}.

Ontology languages have been developed to constrain the topological features of the Semantic Web. The Resource Description Framework Schema (RDFS) supports the representation of subclassing, instantiation, and domain/range restrictions on predicates \cite{rdfspec:manola2004}. The Web Ontology Language (OWL) was developed after RDFS and allows for the creation of more advanced ontologies \cite{owlspec:mcguinness2004}. In OWL, cardinality restrictions, unions, and ontology dependencies were introduced. Semantic Web ontology languages, interestingly, are represented in RDF. Thus, $G$ is the set of all ontologies and their instances.

With RDF, RDFS, and OWL, a medium currently exists to model any physical or conceptual entity and their relationships to one another. The Semantic Web supports universal modeling and allows for the commingling of disparate heterogeneous models within a single substrate that can be used by humans and machines for any computational end. Any statement, logical or illogical, true or false, possible or impossible, can be made explicit in the Semantic Web. While the Semantic Web is primarily used to define descriptive models, there is nothing that prevents the representation of procedural models. In other words, models of computing can be explicitly represented in $G$. It is this modeling power that has prompted the growth of the semantic computing paradigm where the Semantic Web is no longer perceived solely as a universal data modeling medium, but also as a universal computing platform.

While the ideas presented in this article are amenable to any semantic network representation, this article will focus primarily on the Semantic Web due in large part to the technological infrastructure that currently supports this effort. This article's exploration will begin with a review of the various aspects of $G$. Next, a formal definition of computing will be presented in order to describe how the various components of computing can be represented by a semantic network. Current semantic network computing models will be placed within this semantic computing space. The definition of this space will expose areas that have yet to be developed and leave open the potential for future work in the area of semantic network computing.

\section{Descriptive and Procedural Models}

\begin{figure}[h!]
	\centering
	\includegraphics[width=0.48\textwidth]{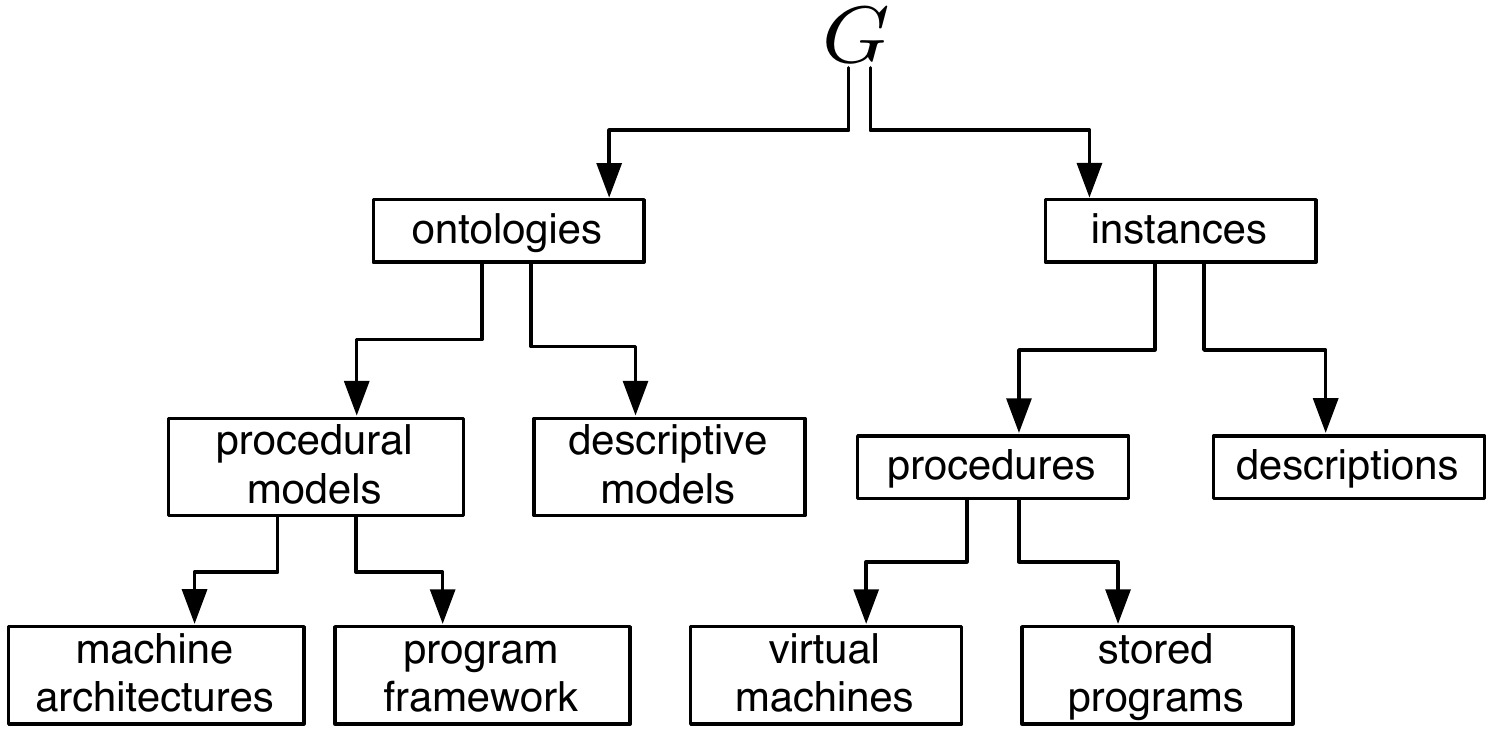}
	 \caption{\label{fig:overview}The descriptive and procedural components of $G$.}
\end{figure}

Currently, the Semantic Web is perceived primarily as a data modeling environment where data is more ``descriptive" rather than ``procedural" in nature \cite{ripple:shinavier2007}. In other words, the triples in $G$ define a model, not the rules by which that model should evolve. This article will explore the more procedural aspects of $G$.  Figure \ref{fig:overview} presents an taxonomy of the various types of triples contained in $G$, where edges have the semantic ``composed of".

In its whole, $G$ is composed of nothing but triples. However, particular subsets of $G$ are used to represent different aspects of the larger $G$ model. Due to RDF, RDFS, and OWL, $G$ is composed of two main subnetworks: the ontological subnetwork and the instance subnetwork. While, in principle, anything can be modeled by a semantic network, most ontologies and instances are descriptive. However, there is nothing that prevents RDF from being used as a framework for denoting procedural models. That is, $G$ can be used to model functions (i.e.~programs) and the machines that execute those functions.

This article will focus on the procedural aspects of $G$. Ontological procedural models represent machine architectures (i.e.~abstract machines) and the abstract functions for which they process. On the other hand, instantiated procedures are stored programs (i.e.~functions, algorithms, etc.) that are explicitly encoded for virtual machines (i.e.~instances of an abstract machine architecture) to execute. The next section will present a formal description of computing.

\section{Representing Computations in a One-Dimensional Tape}

The classic notion of a computation is any process that can be explicitly represented by a formal algorithm. An algorithm is a sequence of executable, well-defined instructions \cite{stone:orgdata1972}. This sequence of instructions is executed by some system, or machine. This machine may contain, internal to it, all the requirements necessary to render the results of the algorithm or, in other instances, may rely on some external storage medium to read in novel inputs and write novel outputs. If the former computing model is chosen, then the machine can only execute a single algorithm with no variation on its behavior because no new input is altering its deterministic path (e.g.~$1 +  2 = 3$). However, if the latter model is chosen, the machine is general-purpose with respects to the particular ``hard-wired" abstract algorithm. It is considered general-purpose because it can map any input to its respective output according to its abstract algorithm (e.g.~$x + y = z$).

This concept can be taken to its logical conclusion where a single machine can be engineered to perform any computing task. Paradoxically, that single machine executes one and only one algorithm. However, that particular algorithm is so generalized, that it can execute any number of other algorithms represented in the machine's external storage medium. This generalized algorithm can reach the ``lowest common denominator" of computing and at that point, can even execute a representation of itself encoded in the storage medium. This machine is called a universal computing machine and is what is know today as the general-purpose computer. This idea was demonstrated by Alan Turing in the 1930s and is the foundation of the computer sciences \cite{turing:herken1994}.

\subsection{Modeling Computations using a Turing Machine}

Perhaps the most common model used to represent computing is the Turing machine \cite{compute:turing1937}. In the Turing machine model of computation, $M$ is a machine with a single read/write head and $D$ is a storage medium called a ``tape" that can be read from and written to by $M$. A Turing machine can be formalized by the 5-tuple 
%%%
\begin{equation*}
M = \la Q, \Gamma, \delta, q_0, d_0 \ra, 
\end{equation*}
%%%
where 
%%%
\begin{itemize}
\item $Q$ is a set of machine states,
\item $\Gamma$ is a set of information symbols (e.g. 0,1),
\item $\delta : Q \times R \rightarrow \{W(\gamma), E\} \times \{lf,rt\} \times Q$ is the transition/behavior function,
\item $q_0 \in Q$ is the start state of the machine,
\item and $d_0 \in D$ is the start location of the machine head on $D$.
\end{itemize}
%%%
$D$ is a one-dimensional $n$-length vector of symbols from $\Gamma$ such that $D  \in \Gamma^n$. 

A Turing machine, $M$, will start at state $q_0 \in Q$ and cell $d_0 \in D$. Depending on what $\gamma \in \Gamma$ is read ($R$) at $d_0$, $M$ will use its $\delta$ function to determine:  1.) what $\gamma \in \Gamma$ to write ($W$) to $d_0$ or whether to erase ($E$) the current symbol, 2.) whether to move its read/write head left ($lf$) or right ($rt$) on $D$, and finally 3.) determine which state in $Q$ to transition to at the next time step,. This 5-tuple model is a simplified version of the 7-tuple representation in \cite{hopcroft:automata1979}. 

Let $M$ denote a Turing machine that increments a unary number by one. While this is not the most exciting algorithm, it is simple enough to represent succinctly and provides an example of the previous abstract concept. The $\delta$-function for $M$ is 

\begin{figure}[h!]
	\centering
	\includegraphics[width=0.3\textwidth]{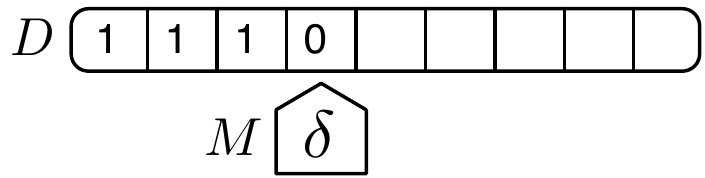}
	 \caption{\label{fig:unary-incrementer}A Turing machine uses the tape for its input and output.}
\end{figure}

\begin{table}[h!]
\begin{footnotesize}
\begin{center}
\begin{tabular}{|c|c||c|c|c|c|}\hline
$q_i$ & $R$ & $W$ & $E$ & move & $q_{i+1}$ \\ \hline\hline
\textbf{A} & 0 & 1 & $\emptyset$ & $rt$ & \textbf{B} \\\hline
\textbf{A} & 1 & $\emptyset$ & $\emptyset$ & $rt$ & \textbf{A} \\\hline
\textbf{B} & $\emptyset$ & $\emptyset$ & $\emptyset$ & $\emptyset$& \textbf{B} \\\hline
\end{tabular}
\end{center}
\end{footnotesize}
\end{table}

where $M$ will write a $1$ if a $0$ exists at its current $d \in D$, else it will move right and replay state \textbf{A} and the state \textbf{B} is considered the halt state. Thus, if $D = (1,1,0,0)$, $M$ will read the first $1$, move right, read the second $1$, move right, read the first $0$, and write a $1$. Upon entering state \textbf{B}, $D = (1,1,1,0)$. At the completion of this algorithm, the number $2$ (11) is incremented to $3$ (111). $M$ and $D$ are represented in Figure \ref{fig:unary-incrementer}.

Imagine having a single physical machine for every computation one required to execute. For instance, one would have an $M$ to add integers, an $M$ to divide floating-points, an $M$ to compare a string of characters, etc. To meet modern computing requirements, an unimaginable number of machines would be required. However, in fact, a single machine does exist for each computing need! Fortunately, these machines need not be physically represented, but instead can be virtually represented in $D$. This is the concept of the stored program and was serendipitously discovered by Alan Turing when he developed the idea of the universal Turing machine \cite{compute:turing1937}.

\subsection{Modeling Computations using a Universal Turing Machine}

\begin{figure}[h!]
	\centering
	\includegraphics[width=0.3\textwidth]{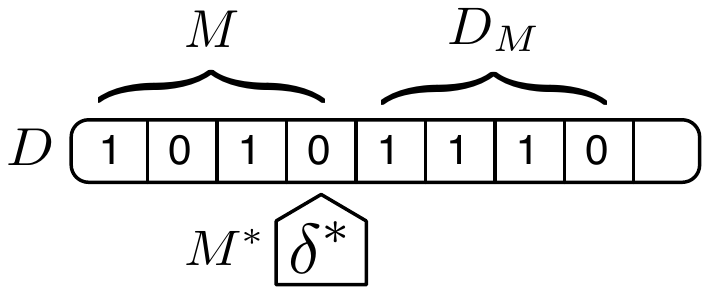}
	 \caption{\label{fig:universal-turing}A universal Turing machine can execute the behavior of any Turing machine.}
\end{figure}

A universal Turing machine, $M^*$, is a Turing machine that can execute the behavior of another Turing machine, $M$. This idea is a central tenet to the engineering of modern day computers. With a universal Turing machine, the state behavior of $M$ can be encoded on $D$ such that some $M^*$ can simulate the behavior of the $M$ encoded in $D$. In such cases, there exists another portion of $D$ that serves as the input/ouput to $M$ denoted $D_M \subset D$. This idea is depicted in Figure \ref{fig:universal-turing}.

The benefit of $M^*$ is that $M^*$ is a general-purpose machine that can be used to execute any algorithm. Thus, there need not exist separate physical machines for each algorithm. However, in order for $M^*$ to execute some $M$, $M$ must be encoded such that it is congruent with the expectations of $M^*$'s $\delta$-function. Thus, there exists an ontology, $\hat{M}$, defining the requirements of the $M$ encoding. If some $M$ is represented according to $\hat{M}$, then $M^*$ can execute it. In the lexicon of modern computing, if a program is written in native machine code, then the native machine can execute it.

\begin{figure}[h!]
	\centering
	\includegraphics[width=0.3\textwidth]{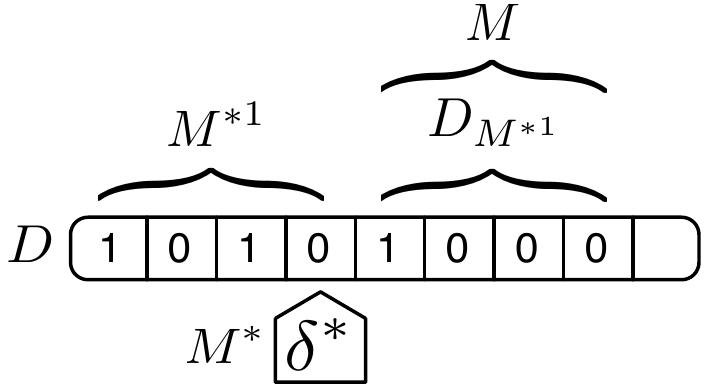}
	 \caption{\label{fig:multi-universal}A universal Turing machine can execute another universal Turing machine that is executing a Turing machine.}
\end{figure}

Finally, to present the conclusion of this chain of reasoning, it is possible for $M^*$ to be encoded according to the $\hat{M}$ ontology. Let $M^{*}$ denote the physical machine and $M^{*1} \subset D$ denote the virtual $D$-encoded machine that is congruent with $\hat{M}$. In such cases, $M^{*1}$ can be used to execute some other $M$ in $D_{M^{*1}} \subset D$. This idea is diagrammed in Figure \ref{fig:multi-universal}. This idea is congruent with the concept of the virtual machine of modern day computing \cite{vm:craig2005}.

\section{Representing Computations in a Semantic Network}

While the Turing model of computing is very simple, it is actually quite representative of the current state of computing in semantic networks. The Semantic Web's $G$ is a data structure similar to $D$ except that $G$ is not a one-dimensional vector of $\Gamma$ symbols. While it is possible to represent $G$ as a one-dimensional string of $\Gamma$ symbols, the more intuitive and useful representation is that of a network of URIs ($U$) and literals ($L$). $G$ is a highly-distributed universal ``tape" that can be accessed by machines world-wide for various computational purposes. However, how much of $G$ is leveraged for computing is machine-instance dependent.

Currently, the actual application that explicitly encodes subsets of $G$ is the triple-store (i.e.~graph database, semantic repository, etc.). A triple-store is a database that contains a subset of the larger Semantic Web. The triple-store is the gatekeeper for determining how triples are read from and written to the Semantic Web. Unlike the read/write head of the Turing machine, the machines that access $G$ are able to move about $G$ in a more random-access fashion due to the development of the common variable-binding interface. While any other $G$ interface may be developed in the future, the lowest-level requirements of such an interface are the ability to read, write, and delete triples from $G$. This section will discuss the nature of a primitive read/write interface into $G$ and its relation to $G$-computing.

\section*{------}

As demonstrated by Alan Turing, the most primitive components required for a computing machine are the ability to read and write to a medium and alter its states according to its perception of that medium. Similar to the relationship between $M$ and $D$, it is possible to develop a semantic Turing machine that is able to read/write to $G$ and evolve its state behavior accordingly.

A semantic Turing machine is denoted $S$ and can be formalized by the 5-tuple
%%%
\begin{equation*}
S = \la Q, \Gamma, \delta, q_0, X \ra,
\end{equation*}
%%%
where
%%%
\begin{itemize}
\item $Q$ is a set of machine states,
\item $\Gamma \subseteq U \cup L$ is the set of URI and literal symbols,
\item $\delta : Q \times R(\varphi) \times X \rightarrow \{W(\tau), E(\varphi)\} \times Q \times Q$ is the transition/behavior function,
\item $q_0 \in Q$ is the start state of the machine,
\item and $X$ is a set of random access machine heads.
\end{itemize}

These components will be discussed in full throughout the remainder of this section. 

The most readily used low-level read model for the Semantic Web is the 3-element symbol binding model,
%%%
\begin{equation*}
R : \varphi \rightarrow \tau \in G,
\end{equation*}
%%%
where $\varphi$ is called a query, $\varphi = \la a, b, c \ra$, and the elements $a$, $b$, and $c$ can either be drawn from the set $\Gamma = U \cup L$ or from the set of machine heads defined by $X$. If those heads in $X$ are declared bindings, then the machine head is random access. In a semantic Turing machine, there does not exist an explicit move behavior. If a state $q \in Q$ is to move a random access head, then it places a bind-symbol before the head name (e.g.~$?x1$) otherwise the machine will hold its head at its current pointed to location with a static-symbol (e.g.~$!x1$). For instance, $R(\la \ttt{marko}, \ttt{isA}, ?x1 \ra)$ would place the head $?x1$ on some object of a triple with the subject \ttt{marko} and predicate \ttt{isA}. If $?x1$ bound to $\ttt{human}$ then $\tau = \la \ttt{marko}, \ttt{isA}, \ttt{human} \ra \in G$. However, if the machine head is already at a particular resource in $G$, then it can be used as a static variable. If $?x1$ bound to \ttt{human} on a previous read, then $R(\la !x1, \ttt{subClassOf}, ?x2 \ra)$ will move $?x2$ to the resource \ttt{mammal}. With the random-access $X$ machine heads, no variable states are represented internal to $S$, they are simply pointed to by some $x \in X$ in $G$. 

The most readily used write model for the Semantic Web is to union the semantic network triple list $G$ with a new triple $\tau$,
%%%
\begin{equation*}
W : \tau \rightarrow G \cup \tau,
\end{equation*}
%%%
where $\tau = \la s,p,o \ra$, $s \in U$, $p \in U$, and $o \in (U \cup L)$.

Finally, in order to erase (i.e.~delete) a triple, the 3-element symbol binding model can be used,
%%%
\begin{equation*}
E : \varphi  \rightarrow G \setminus R(\varphi),
\end{equation*}
%%%
where the triple $R(\varphi) \in G$ is removed from $G$.

An $S$ can be built to do any type of computation on $G$. The popular Horn-clause query/assertion can be represented by an $S$ \cite{clause:horn1951}. For instance, the rule 
%%%
\begin{align*}
& \ttt{hasParent}(\ttt{marko},?x1) \; \wedge \; \ttt{hasBrother}(?x1,?x2) \\
& \;\;\;\;\; \rightarrow \ttt{hasUncle}(\ttt{marko},?x2) 
\end{align*}
%%%
states that if \ttt{marko} has a parent that binds to $?x1$ and $x1$ has a brother that binds to $?x2$ then assert (i.e.~write) the fact that $x2$ is \ttt{marko}'s uncle. The $\delta$-function for $S$ that executes this query is

\begin{table}[h!]
\begin{footnotesize}
\begin{center}
\begin{tabular}{|c|c|c||c|c|c|c|}\hline
$q_i$ & $R$ & $X$ & $W$ & $E$ & $q_{i+1}$ & $\not\varphi$ \\ \hline\hline
\textbf{A} & $\la \ttt{marko}, \ttt{hasParent}, ?x1 \ra$ & $\emptyset$ & $\emptyset$ & $\emptyset$ & \textbf{B} & \textbf{C} \\\hline
\textbf{B} & $\la !x1, \ttt{hasBrother}, ?x2 \ra$ & $\emptyset$ & $\la \ttt{marko}, \ttt{hasUncle}, !x2 \ra$ & $\emptyset$ & \textbf{C} & \textbf{C} \\\hline
\textbf{C} & $\emptyset$ & $\emptyset$ & $\emptyset$ & $\emptyset$ & \textbf{C} & $\emptyset$ \\\hline
\end{tabular}
\end{center}
\end{footnotesize}
\end{table}

where $q_0 = \textbf{A}$, \textbf{C} is the halt state, $x1, x2 \in X$ and $\not\varphi$ is the state transition when a $\varphi$ fails. If
%%%
\begin{align*}
G=&\{\la \ttt{marko}, \ttt{hasParent}, \ttt{carole} \ra, \\
     & \; \; \la \ttt{carole}, \ttt{hasBrother}, \ttt{george} \ra \},
\end{align*}
%%%
then at $q_0 =  \textbf{A}$, $?x1$ will point to \ttt{carole}, at $q_1 =  \textbf{B}$, $?x2$ will point to \ttt{george}, and at $q_3 =  \textbf{C}$,
%%%
\begin{align*}
G=&\{\la \ttt{marko}, \ttt{hasParent}, \ttt{carole} \ra, \\
     & \; \; \la \ttt{carole}, \ttt{hasBrother}, \ttt{george} \ra \\
     & \; \; \la \ttt{marko}, \ttt{hasUncle}, \ttt{george} \ra \}.
\end{align*}

For more arithmetic operations and for the construction of novel URIs and literals, the classic Turing machine model can be used for writing triples that bind symbols in a list-like fashion. In other words, a semantic network can simulate a one-dimensional tape. In this model, the semantic Turing machine utilizes only $G$ for its workspace computations and the semantic Turing machine is analogous in terms of its component parts to the classic Turing machine. The $\delta$-function to increment a unary number by $1$ is
%%%
\begin{table}[h!]
\begin{footnotesize}
\begin{center}
\begin{tabular}{|c|c|c||c|c|c|c|}\hline
$q_i$ & $R$ & $X$ & $W$ & $E$ & $q_{i+1}$ & $\not\varphi$ \\ \hline\hline
\textbf{A} & $\la \ttt{bit1}, \ttt{hasValue}, ?x1 \ra$ & x1 = 0 & $\la \ttt{bit1}, \ttt{hasValue}, 1 \ra$ & $\emptyset$ & \textbf{F} & \textbf{F} \\\hline
\textbf{A} & $\la \ttt{bit1}, \ttt{hasValue}, ?x1 \ra$ & x1 = 1 & $\emptyset$ & $\emptyset$ & \textbf{B} & \textbf{F} \\\hline
\textbf{B} & $\la \ttt{bit1}, \ttt{nextBit}, ?x2 \ra$ & $\emptyset$ & $\emptyset$ & $\emptyset$ & \textbf{C} & \textbf{F} \\\hline
\textbf{C} & $\la !x2, \ttt{hasValue}, ?x3 \ra$ & x3 = 0 & $\la !x2, \ttt{hasValue}, 1 \ra$ & $\emptyset$ & \textbf{F} & \textbf{F} \\\hline
\textbf{C} & $\la !x2, \ttt{hasValue}, ?x3 \ra$ & x3 = 1 & $\emptyset$ & $\emptyset$ & \textbf{D} & \textbf{F} \\\hline
\textbf{D} & $\la !x2, \ttt{nextBit}, ?x4 \ra$ & $\emptyset$ & $\emptyset$ & $\emptyset$ & \textbf{E} & \textbf{F} \\\hline
\textbf{E} & $\la ?x2, \ttt{nextBit}, !x4 \ra$ & $\emptyset$ & $\emptyset$ & $\emptyset$ & \textbf{C} & \textbf{F} \\\hline
\textbf{F} & $\emptyset$ & $\emptyset$ & $\emptyset$ & $\emptyset$ & \textbf{F} & $\emptyset$ \\\hline
\end{tabular}
\end{center}
\end{footnotesize}
\end{table}

where the URI \ttt{bit1} is the subject of the triple whose object is the first bit of the unary number. While it is possible to perform low-level arithmetic calculations in $G$, constructing such a machine is impractical. Unlike a physical $M$ where the laws of physics are the computing substrate, $S$s are embedded in a substrate that was engineered for computing--the general-purpose processor. Thus, an $S$ can rely on its local processor for arithmetic computations and for the construction of new URIs and literals. What was presented previous was only to demonstrate that $G$ can be used as a universal computing ``tape". However, how much of computing is represented in $G$ is implementation specific, but the more a computation is represented in $G$, the more the Semantic Web can be made to behave like a general-purpose computer.

Finally, it is possible to represent the previous two $\delta$-functions in $G$ such that some $S^*$ external to $G$ is able to simulate the behavior of these respective $S$ machines. In this sense, $S^*$ is a universal semantic Turing machine and any $S \subset G$ that obeys the $\hat{S}$ ontology can be executed by $S^*$. The next section will discuss moving computations into $G$ to ultimately arrive at a general-purpose computer embedded in $G$--a semantic virtual machine.

\section{The Semantic Web as a General-Purpose Computer}

The current state of the Semantic Web is such that machines (i.e.~processes) exist external to $G$ and manipulate $G$ by reading, writing, and deleting triples to and from it. In many cases, $G$ does not encode stored programs in the Turing sense. Those processes that manipulate $G$ use some other $D$-medium (e.g.~local memory) for their respective calculations. However, by leveraging external $D$-mediums that are not $G$, there exist multiple machines (i.e.~software programs) that do very specific computing tasks. This is analogous to having different physical $M$s for each desired computing task. On the other hand, when $G$ is leveraged as the sole substrate for encoding information, then it is possible to not only use $G$ for stored programs, but also to use $G$ to represent a universal computing machine \cite{rodriguez:gpsemnet2007}. The benefit of this latter model is that the Semantic Web becomes a universal computing platform, where any number of universal computing machines exist external to $G$ executing the state evolution of those $G$ encoded machines. At this stage, $G$ is a massive computer distributed across servers world-wide. 

The remainder of this section will present the various levels of machine encodings currently realized by the Semantic Web community. The first is the ``external program" level where $S$ machines are external processes whose stored programs are represented in some other $D$-medium. The second level is the ``stored program" level where $S$ machines are external process whose stored programs are represented in $G$. The final level is called the ``virtualized machine" level where $S$ machines are internal processes represented in $G$ whose stored programs are also represented in $G$. 

\subsection{The External Program Model}

In the external program model, the Semantic Web is considered a database. The machines (i.e.~programs) developed for $G$ exist external to $G$ and only use $G$ for reading descriptive data (and possibly writing descriptive data). This is analogous to the physical manifestation of an algorithm in the Turing model of computing. While it is possible for $S$ to be completely configurable and thus, not ``hard coded", this $\delta$-function is stored in a separate $D$-medium where $G \cap D = \emptyset$. Therefore, with respects to $G$ as a general computing platform, this model is the farthest removed from this vision.

\begin{figure}[h!]
	\centering
	\includegraphics[width=0.3\textwidth]{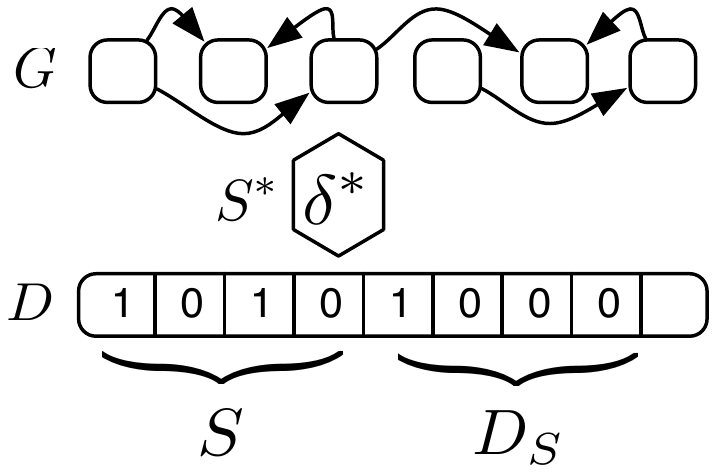}
	 \caption{\label{fig:d-sparql}A $D$-encoded machine reads its input from $G$.}
\end{figure}

One such example of the external program model is the SPARQL query language \cite{sparql:prud2004}. The SPARQL query language is a Horn-based \cite{clause:horn1951} query language that supports semantic searching in $G$. For instance, the following example SPARQL query
%%%
\begin{verbatim}
SELECT ?x 
WHERE { marko isA ?x . }
\end{verbatim}
%%%
will bind $?x$ to all URIs that are the object of a triple that has \ttt{marko} as the subject and \ttt{isA} as the predicate. While a universal SPARQL machine, denoted $S^*$, can execute any SPARQL-proper query, these queries (i.e.~programs) are not explicitly represented in $G$, but instead in some $D$. Thus, with respects to $G$, each SPARQL query is analogous to a unique $S$. Furthermore, the output from any $S$ is encoded in $D$ (more specifically $D_S$). In this sense, $G$ is only used as the input parameter to $S$, not as a computational ``workspace". This computing model is diagrammed in Figure \ref{fig:d-sparql}, where $S^*$ is the universal SPARQL machine, $S$ is a particular SPARQL query, and $D_S$ is the result set derived from the execution of $S$ on $G$.

It is no large conceptual leap to actually encode SPARQL queries in RDF and therefore, in $G$. In fact, the semantic network data structure is an ideal medium for many types of information encodings due to its generalized network nature that naturally supports the expression of trees, lists, graphs, tables, etc. The next subsection will discuss such stored programs.

\subsection{The Stored Program Model}

In the stored program model, $S^*$ is a universal machine that reads its parameter specification from $G$ and writes its algorithm's output to $G$. Thus, the specification of the $\delta$-function of any $S$ is encoded in $G$.

\begin{figure}[h!]
	\centering
	\includegraphics[width=0.3\textwidth]{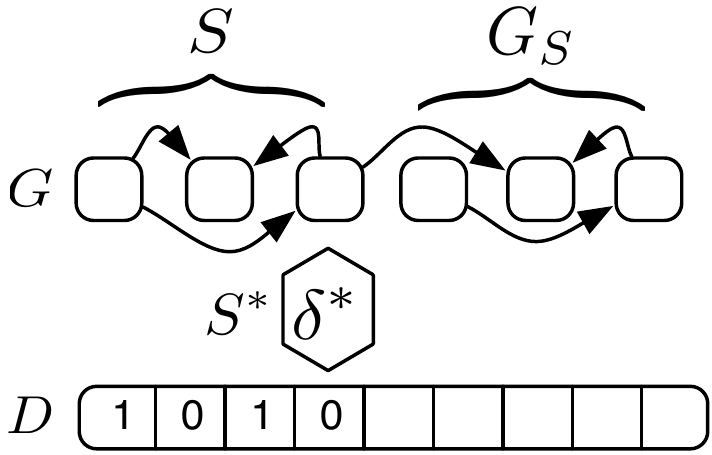}
	 \caption{\label{fig:universal-swrl}A universal $S^*$ can execute any $S$ in $G$.}
\end{figure}

One such example of the stored program model is the Semantic Web Rule Language (SWRL). SWRL is a Horn-clause based query/assertion language similar to SPARQL \cite{swrl:horrocks2004}. For example, in the ``my friend is your friend" query/assertion
%%%
\begin{align*}
& \ttt{hasFriend}(?x1,?x2) \; \wedge \; \ttt{hasFriend}(?x2,?x3) \\
& \;\;\;\;\; \rightarrow \ttt{hasFriend}(?x1,?x3)
\end{align*}
%%%
if $?x1$ has a friend $?x2$ and $?x2$ has a friend $?x3$, then $?x1$ and $?x3$ are asserted to be friends. Interestingly, SWRL query/assertions can be represented in RDF and thus, can be explicitly encoded in $G$. The benefit of this is that there can exist a generalized SWRL machine denoted $S^*$ that can point to any particular $S$ in $G$. This idea is depicted in Figure \ref{fig:universal-swrl}, where $S^*$ is a universal SWRL machine, $S$ is a particular SWRL query/assertion, and $G_S$ is the result of the execution of $S$. However, note that $D$ is the computational workspace for $S^*$, not $G$.

While SWRL $S^*$ is Turing complete \cite{compute:brainerd1974}, it is impractical to represent arithmetic and loop-based algorithms in SWRL. This means that there exists algorithms for which SWRL $S^*$ can not easily emulate. As a remedy to this problem, two Turing complete $S^*$ machines were developed: the stack-based Ripple \cite{ripple:shinavier2007} and the RAM-based r-Fhat \cite{rodriguez:gpsemnet2007}. For Ripple and r-Fhat, like SWRL, their respective programs are encoded in $G$. However, both Ripple and r-Fhat maintain their respective universal machine data structures in $D$ for computing local operations and thus, do not completely use $G$ as their computing workspace.

In the stored program model, there not only exists descriptive data in $G$, but also procedural data. In many cases, there also exists an ontology $\hat{S}$ that defines the structure of that procedural data. In general, if a subset of $G$ obeys $\hat{S}$, then it is computable by $S^*$. The next subsection will discuss full machine virtualization and the explicit representation of $S^* \subset G$.

\subsection{The Virtualized Machine Model}

The previous section discussed the explicit encoding of stored programs in $G$. However, there is nothing preventing the stored program from being a computing machine. In this model, a virtual machine is encoded in $G$ along with the programs that the virtual machine executes. In order to represent a virtual machine in $G$ is it necessary to support a write/delete interface to $G$ since the machine evolution and its effect on $G$ is the computation.

\begin{figure}[h!]
	\centering
	\includegraphics[width=0.3\textwidth]{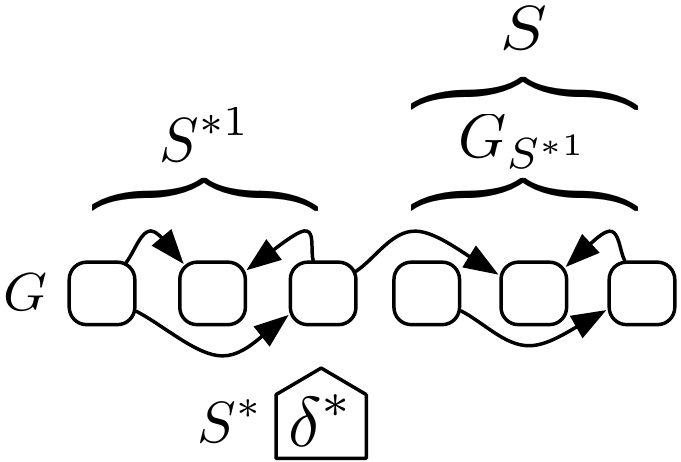}
	 \caption{\label{fig:universal-fhat}A universal $S^*$ can execute a universal $S^{*1}$, which is executing an $S$.}
\end{figure}

Currently, the only example of a virtualized machine encoded in $G$ is the Fhat RVM (RDF virtual machine) \cite{rodriguez:gpsemnet2007}. A Fhat processor, denoted $S^{*1}$ exists internal to $G$. Another process $S^*$ external to $G$ is a general-purpose machine that reads $S^{*1}$ from $G$ as if it were any other program. However, $S^{*1}$ is not only a program, but is another machine that is executing an algorithm, $S$, in another area of $G$, $G_{S^{*1}}$. The virtualized machine model is depicted in Figure \ref{fig:universal-fhat}.

There is nothing that prevents the $S$ that $S^{*1}$ is executing from being another $S^*$. For example, imagine two $S^*$ machines encoded in $G$: $S^{*1}$ and $S^{*2}$, where $S^{*1} \cap S^{*2} = \emptyset$. The external $S^*$ can be reading in $S^{*1}$ as a program, which is reading in $S^{*2}$ as a program, which is reading in some other machine $S$ as a program. In this model, there is no limit to the amount of computing redirection that is possible. Ultimately, it is up to the external $S^*$ to perform all the read/write operations that update the respective states of all the chained together $S^{*n}$ machines.

In the virtual machine model, not only is procedural data encoded in $G$, but also machine data. There must exist both an ontology for procedural data $\hat{S}$ and an ontology for machine data $\hat{S^*}$. In principle, any subset of $G$ that obeys $\hat{S^*}$ is a virtualized computing machine.

\subsection{True Universality}

While a universal semantic Turing machine can be created, it is impractical to do so because of the speed constraints currently realized by the read/write interface to the Semantic Web and because any external $S^*$ already exists in a substrate engineered for general-purpose computing. Therefore, for virtualized semantic machines, $D$-mediums are currently used for low-level arithmetic operations only. There will always be a tradeoff between the desire to represent low-level computations in $G$ and the desire to ensure the fast execution of those $G$-based machine representations. 

The Fhat processor was designed with this constraint in mind. Many aspects of the machine's state are represented in $G$ including its operand stack, symbol table, program counter, etc. However, when a low-level operation such as \ttt{add 2 3} is called, those values are calculated on the physical machine, not in $G$. While this may not be completely theoretically satisfying, it does support a practical implementation of the virtual machine model of computing in $G$.

\section{The Future of Semantic Network Computing}

The future of semantic network computing may be one in which virtual machines and their programs exist in $G$. Any universal machine external to $G$ can gain access to the URI denoting a virtual machine and begin to execute its ``physics". In other words, evolve its state and compute. In this idealized world, the underlying physical hardware supporting the execution of these virtual machines is more or less inconsequential. These underlying hardware processors are analogous to the underlying physics supporting the execution of our hardware machines. Once the protocols are in place to ensure that $G$ has a farm of processors continuously evolving it, then the Semantic Web will have reached a transition to where abstract virtualized computing becomes ubiquitous and $G$ can be seen as a single distributed computer with the massive address space of $U \cup L$. However, there are still many obstacles that prevent this model from becoming a common reality.

First, the read/write speeds for $G$ are orders of magnitude slower than the read/write speeds for local memory and thus, computing in $G$ is orders of magnitude slower. There is still much more room for growth in the area of triple-store index algorithms. Unlike the relational database model where data is broken into different linked tables, the triple-store is a single massive table with various indexes supporting fast searching. As the read/write speeds continue to increase, the ability to use $G$ as a computing ``tape" will become more viable.

Third, current triple-store's have limits on the number of triples they can feasibly represent in a single store. While some stores can easily support up to $10^9$ triples, the explicit representation of procedural data reduces the amount of space available for descriptive data. Fortunately, with an increase in the use of standards liked Linked Data \cite{berners:ldata2006}, the growth of $G$ will have limited effect on the ability to compute in $G$.

Fourth, the current state of affairs in the Semantic Web is such that writing to $G$ is cumbersome due to the absence of a generally accepted protocol to do so. While the proposed SPARQL/Update protocol \cite{sparqlupdate:seaborne2007} is one such write interface, it is not widely supported by all triple-store providers. Thus, each triple-store provider maintains their own mechanism for writing and deleting triples.

Finally, there does not exist a universal trust and security mechanism to deter malicious machines in $G$. If $G$ is conceived as a a universal computing ``tape", then the read, and more importantly, write/delete accesses to $G$ will need to be established. Of course, $G$ is only contained in an abstract universal store. Each triple-store supports only a subset of the larger whole. Therefore, for those running a triple-store, read/write privileges is not an issue. However, as more procedural information is encoded in $G$ and machines can share procedural fragments, understanding where particular bits of information were derived from becomes very important. Work in the area of named graphs for trust and provenance should prove promising in this area \cite{named:carroll2005}. The named graph extends the triple concept by adding an extra resource called $g$, or graph. A triple is thus a quad and $\tau = \la s, p, o, g \ra$. The $g$ component of $\tau$ is a URI and this information can be used to attach read/write privileges to particular subnetworks of $G$.

While this list is not conclusive, it provides an overview of some of the more prominent issues concerning the future of semantic network computing. 

\section{Conclusion}

Given that the Semantic Web is an abstract data structure, it does not have the capacity to perform a computation in and of itself. The Semantic Web is simply a description of the relationship between URIs and literals and, in order to evolve, it requires the explicit contribution of external machines to read and write to it. However, the amount of procedural information that is actually encoded in the Semantic Web can vary. At one extreme, the Semantic Web is a read-only substrate that has limited effect on how a computation evolves. At the other extreme, the Semantic Web is the representational substrate for not only the data aspects of a computation, but also the algorithmic and machine representations as well.

This article has presented an analysis of the various models of computing in the Semantic Web and in semantic networks in general. It is the hope that more research and development will go into developing practical computing environments that leverage $G$ as their computing substrate.

\section*{Acknowledgments}

Marko A. Rodriguez and Johan Bollen are funded by a grant from the Andrew W. Mellon Foundation.

\end{document}